\documentclass[preprint]{elsarticle}

\usepackage{amssymb}
\usepackage{color}
\usepackage{graphicx,amssymb,bm}

\usepackage{float}
\usepackage{algorithm}
\usepackage{algorithmic}
\usepackage{booktabs}
\usepackage{threeparttable}
\usepackage[a4paper, body={18cm,24cm}]{geometry} 
\usepackage{amsmath}
\usepackage{color}
\usepackage[usenames, dvipsnames]{xcolor}
\usepackage{mathrsfs}
\usepackage[colorlinks]{hyperref}

\usepackage{enumerate}
\usepackage{relsize}

\usepackage{multirow}
\usepackage{lineno}

\journal{Neurocomputing}

\begin{document}
% \linenumbers

\begin{frontmatter}

\title{Positive-Negative Equal Contrastive Loss for Semantic Segmentation}

\author{Jing Wang$^{1}$}\ead{m202120718@xs.ustb.edu.cn}

\author{Jiangyun Li$^{1}$\corref{cor1}}\cortext[cor1]{Corresponding author}\ead{leejy@ustb.edu.cn}

\author{Wei Li$^{1}$}

\author{Lingfei Xuan$^{1}$}

\author{Tianxiang Zhang$^{1}$}\ead{txzhang@ustb.edu.cn}

\author{Wenxuan Wang$^{1}$ \\ $^1$University of Science and Technology Beijing}

\begin{abstract}
  The contextual information is critical for various computer vision tasks, previous works commonly design plug-and-play modules and structural losses to effectively extract and aggregate the global context. These methods utilize fine-label to optimize the model but ignore that fine-trained features are also precious training resources, which can introduce preferable distribution to hard pixels (i.e., misclassified pixels). Inspired by contrastive learning in unsupervised paradigm, we apply the contrastive loss in a supervised manner and re-design the loss function to cast off the stereotype of unsupervised learning (e.g., imbalance of positives and negatives, confusion of anchors computing). To this end, we propose \textbf{P}ositive-\textbf{N}egative \textbf{E}qual contrastive loss (PNE loss), which increases the latent impact of positive embedding on the anchor and treats the positive as well as negative sample pairs equally. The PNE loss can be directly plugged right into existing semantic segmentation frameworks and leads to excellent performance with neglectable extra computational costs. We utilize a number of classic segmentation methods (e.g., DeepLabV3, HRNetV2, OCRNet, UperNet) and backbone (e.g., ResNet, HRNet, Swin Transformer) to conduct comprehensive experiments and achieve state-of-the-art performance on three benchmark datasets (e.g., Cityscapes,  COCO-Stuff and ADE20K). Our code will be publicly available soon.
\end{abstract}


\end{frontmatter}

\section{Introduction}

\label{section1}
Semantic segmentation, which aims to classify each pixel in the visual area, is employed in many practical scenes (e.g., autonomous driving and clinical assistance). As an essential task in exploring semantic representation for scene understanding, 
 segmentation is driven by pixel-level labeled data and ingenious segmentation networks \cite{long2015fully, fu2019dual,wang2018non,2018DeepLab, yu2015multi,he2015spatial, liu2021sg, wang2022learning}.

For dense prediction tasks like semantic segmentation, both local and global context are critical.
Based on Fully Convolutional Network (FCN) \cite{long2015fully}, several works explore module-based methods to exploit or aggregate contextual information from the visual representation of pixel embeddings, e.g., pixel-to-pixel attention mechanism \cite{fu2019dual, wang2018non}, multi-layer dilated convolution \cite{2018DeepLab, yu2015multi} and pyramid pooling operation \cite{he2015spatial}. However, these proposed frameworks include two-fold shortcomings. Firstly, the process of extracting informative context in \cite{fu2019dual,wang2018non,2018DeepLab, yu2015multi,he2015spatial} is a passive pattern without any supervision. Secondly, complex relationships between large amount of pixels in an image lead to unbearable computational overheads. Thus, these methods \cite{fu2019dual, wang2018non} can not achieve the trade-off between speed and accuracy.
Furthermore, these networks commonly use the pixel-wise cross entropy loss function and lack supervised attention on contextual information that contains abundant representation of explicit relationship between pixels. 

\begin{figure}{}
    \centering
    \includegraphics[width=0.65\textwidth]{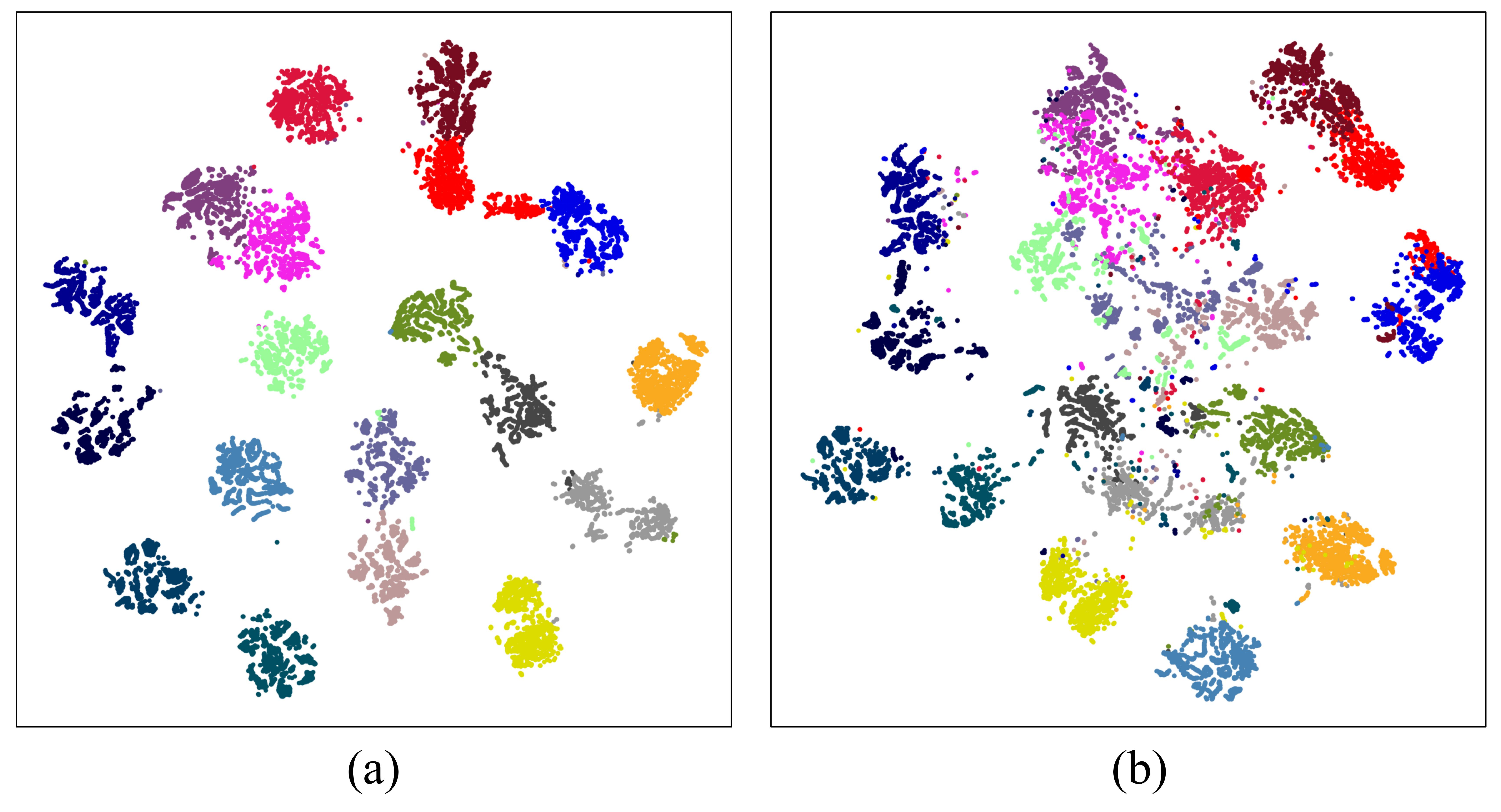}
    \caption{The visualization of pixel embedding on penultimate layer out by t-SNE. As seen, (a) is only sampled correctly classified pixels, and has a more aggregated embedding space. But (b) is sampled by label without concerned about predictions, and shows a confused distribution.}
    \label{Figure.2}
\end{figure}

Some other works explore loss-based methods \cite{ke2018adaptive, zhao2019region} which apply the loss function to guide the network to model relationship between pixels precisely.
These research have suggested that constructing intra- and inter-class structured losses can strengthen pixel-to-pixel interactions.
To illustrate the importance of this inspiration, we visualize the pixel embeddings of the penultimate layer output. As shown in Fig.\ref{Figure.2}, the embedding of correctly classified pixels is more aggregated in embedding space. However, the embedding of wrongly classified pixels, which are generally difficult pixels with semantic confusion, is not uniformly distributed in the expected segmentation embedding space. 
To solve this problem, an intuitive belief is that more consistent the distribution of difficult pixels and correctly classified pixels is, the higher the probability of hard pixels is correctly classified. 
Therefore, designing the loss function to optimize the distribution of pixel embedding makes these confused features close to the embedding of precisely classified pixels, which may be a novel way to assist the network to capture contextual information.

Recently, success in self-supervised learning \cite{chen2020simple,he2020momentum} reveal that contrastive learning shows powerful potential by pre-training model without any labeled data. Contrastive learning force model to extract close feature representations from similar samples (positive pairs) and distant feature embeddings from disparate samples (negative pairs). Each image can be reckoned as an anchor, and the positive pair is achieved through the introduced data augmentation on the current image. At the same time, the negative pair is also constructed from the acquired different image samples. 
This unsupervised paradigm makes the pre-training model show outstanding performance on some specific downstream tasks like image classification and object detection.
Although this aforementioned unsupervised paradigm achieves promising results, it is difficult to construct appropriate positive samples among the large amount of unlabeled data.
Therefore, unsupervised contrastive learning takes advantage of plenty of negative samples which are selected from the acquired datasets. In contrast, supervised semantic segmentation with detailed labeled data can better be incorporated with contrastive learning. Last year, \cite{wang2021exploring, hu2021region} extend the computational domain from intra-image to inter-image, 
aiming to gain more global context from positive and negative samples in the training dataset. With the potential of contrastive learning, they all achieve state-of-the-art performance on semantic segmentation tasks. 

However, such a training paradigm does not effectively integrate the supervised information, which remains two potential drawbacks. \textbf{First}, negative sample pairs instead of positive sample pairs contribute the dominant gradient during parameter optimization so that the attractive force of the positive sample to anchor is not strong enough, which have demonstrated in a previous study \cite{wang2021understanding}.   The value of the original contrastive loss is hardly influenced by positive samples.  
\textbf{Second}, many irrationalities exist in the construction strategy of positive and negative samples in the previous method\cite{wang2021exploring, hu2021region}. For example, both correctly classified and misclassified pixels are utilized as anchors to indiscriminately compute the contrastive loss, anchors with different prediction results share negative samples, and so on. This will lead to problems such as waste of computing resources and confusion in the direction of anchor optimization.
The reason for the above two cases is the limitation of self-supervised contrastive learning. It is a continuation of the lack of annotation and image-level sample expediency in self-supervised learning. Such settings are not the best design for the segmentation task in a supervised way, since it provides pixel-level samples, dense predictions, and accurate labels. Those are valuable information worthy of being fully utilized through a novel dense contrastive loss function.

\begin{figure}[htpb]
\centering
\includegraphics[width=0.48\textwidth]{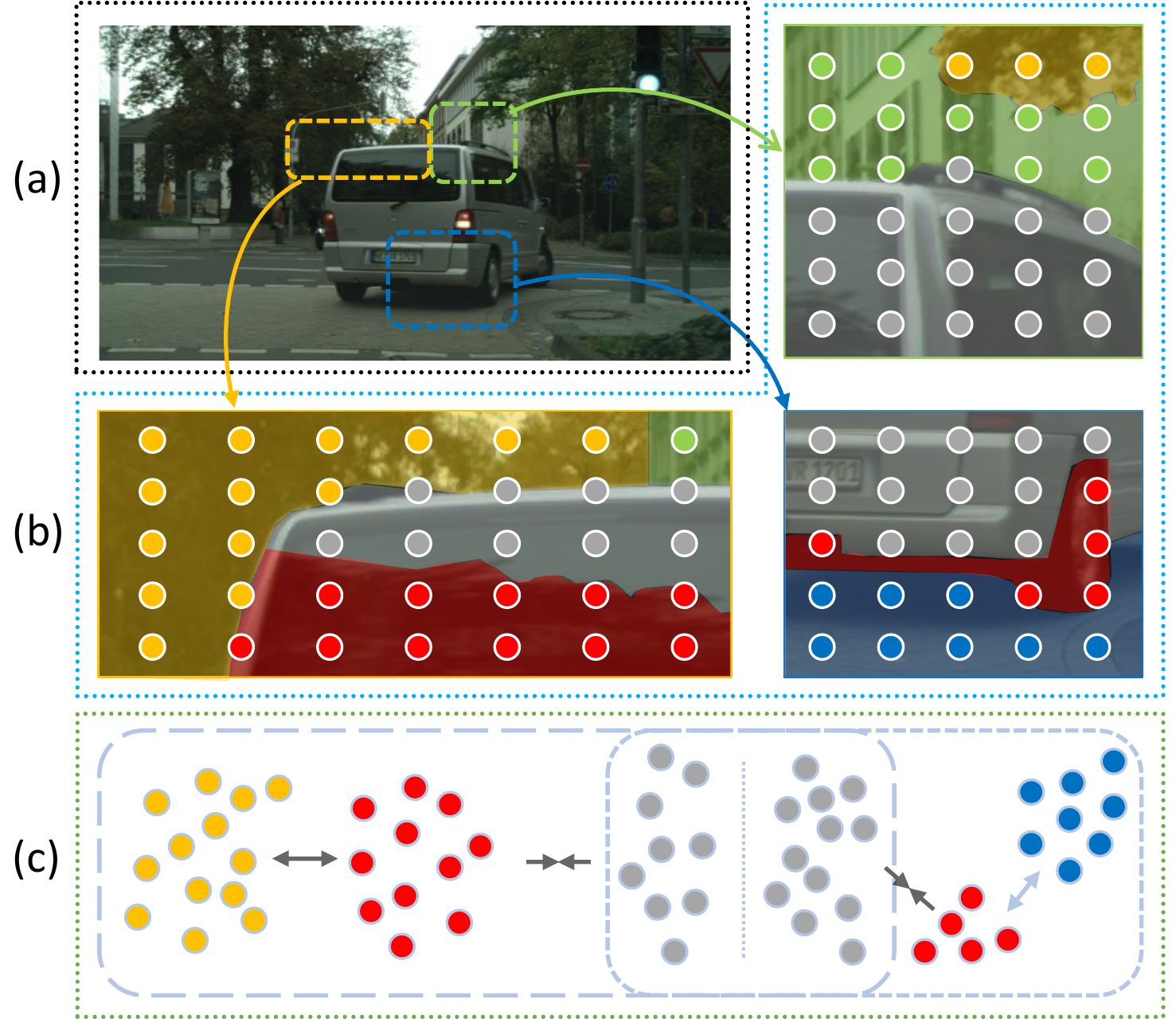}
\caption{\textbf{Main idea}. The segmentation model obtains inference label (b) and pixel embedding (c) from the input image (a). We set the misclassified pixels (difficult samples) in the scene as anchors according to ground-truth (GT). For example, in b-1, the class of the red dots is the car (grey dots) but it is wrongly classified as a tree (yellow dots) because of the reflection of the glass. According to GT, we set red dots as anchors, grey dots as positive samples, yellow dots as negative samples, and building (green dots) with almost no overlap do not participate in the calculation of the contrastive pairs. Additionally, positive samples are collected from all correctly classified pixels, and negative samples are collected only from relevant pixels.}
\label{Figure.1}
\end{figure}

With the above concerns, we propose a novel contrastive loss (PNE loss), named Positive-Negative Equal loss, to supervise pixel-wise embedding by prior knowledge from fine labels. As the common setting, any pixel-wise embeddings extracted by network can be reckoned as a sample. Specifically, positive pairs are constituted with any embedding which corresponds to the same category, while negative pairs are constituted in different categories. There are totally two improvements to our proposed loss function.
\textbf{First}, different from previous work, we modify the calculation method of the loss function to more focus on the guidance from positive samples. It is easy to acquire plenty of positive pairs with finer embedding from pixels in the same class. These positive samples are the best learning templates for anchors with confusing features. As for all the sample pairs, positive and negative have equal importance in the loss function. An anchor point will calculate cosine distance with positive samples repeatedly in a single operation which is just the same as negative pairs, thereby moving toward more general embedding space. \textbf{Second}, we propose a more reasonable pairing method of positive and negative sample pairs, which reduces the amount of computations and makes the optimization direction of the anchor clearer (Fig.\ref{Figure.1}(b)). Only samples which are classified falsely will participate in the calculation as anchors (Fig.\ref{Figure.1}(c)). Furthermore, we prove that contrastive loss is more suitable for the extraction of contextual information in high-level embedding space.

The main contributions of this paper are as follows:
\begin{itemize}
    \item {We propose a novel pixel-wise contrastive loss in a supervised manner that pays more attention to positive pairs that are ignored in previous works. It changes the contrastive learning paradigm from learning target feature to learning inter-class feature embedding space.}
    \item {We design an effective sampling strategy to cast off the confusion of anchors and negatives. 
    Through the decoupling of negative pairs, PNE loss can calculate the semantic association of negative sample pairs more efficiently and reduce the influence of the weak association samples on the anchor.}
    \item {We verify the effectiveness of the proposed PNE loss by conducting extensive experiments with several semantic segmentation networks such as DeepLabV3, OCRNet, and UperNet on various public datasets (i.e., Cityscapes, COCO Stuff, and ADE20K).}
\end{itemize}

\section{Related Works}
\label{gen_inst}
\subsection{Semantic Segmentation}
FCN \cite{long2015fully} uses the fully convolutional network without the full connection layer to achieve the pixel-level prediction task. However, it can only perceive the limited visual context with the local receptive fields, and poorly extract the strong inter-relationship among pixels in an image. Therefore, A lot of research has attempted to enlarge the local sensing domain and enhance the fusion of contextual information for preferably extracting the inter-relationship among pixels. For instance, DeepLab Series \cite{2018DeepLab,chen2017rethinking} employ atrous convolution to increase the receptive fields while ensuring the resolution of the feature map. PSPNet \cite{2016Pyramid} applies multi-scale pooling of the feature pyramids to fuse multi-scale features. U-Net \cite{2015U} and SegNet \cite{2017SegNet} adopt the Encoder-Decoder architecture to merge features from different network layers and recover lost information from downsampling. DANet \cite{fu2019dual} and OCNet \cite{2021OCNet} employ self-attention mechanism to capture long-range context and directly exchange the contextual information between paired pixels. OCRNet \cite{yuan2020object} proposes object-contextual representations that characterize a pixel by exploiting the representation of the corresponding object class. In SETR \cite{zheng2021rethinking} and Segmenter \cite{strudel2021segmenter}, Vision Transformer \cite{dosovitskiy2020image} substitutes traditional convolutional backbones to promote context learning on the lower-level feature by self-attention mechanism. 
    
It is clear that these works focus on module design which is poorly initiative in exploring pixel relation. We propose PNE loss to establish the inherent connection between difficult pixels and other strongly correlated pixels, and provide appropriate optimization direction for difficult pixels.

\subsection{Contrastive Learning}
In self-supervised learning, contrastive learning \cite{oord2018representation,wu2018unsupervised} methods achieve the state-of-the-art result, becoming a major branch in computer vision pretext tasks. It is a discriminative learning manner that contrasts positive (similar) pairs against negative (dissimilar) pairs to learning representations. A vast range of methods based on contrastive manner has been recently proposed. SimCLR \cite{chen2020simple,chen2020big} proposes a simple framework, that constructs the augmented views of the image as positive pairs and the rest of image in the training batch as negative pairs, using huge training batch to exploit many negative pairs efficiently and improve performance. \cite{wu2018unsupervised,misra2020self} use a memory bank to store features of every instance in the dataset, and partly solve the inverse optimization due to the large batch size. MoCo \cite{he2020momentum,chen2020improved} design a momentum encoder to dynamically update negative pairs and maintain consistent representations of negative pairs per iteration. To improve the optimization efficiency, some works \cite{kalantidis2020hard,khosla2020supervised,robinson2020contrastive} sample negative pairs according to their cosine similarity. Moreover, \cite{wang2021understanding} probe into the function of temperature hyper-parameter and find it domain the attention of negative pairs in the loss function.
    
Previous years, some works \cite{wang2021exploring,hu2021region,2020Contrastive1,wang2021dense,2020Propagate1}, have adopted pixel-to-pixel contrastive loss and achieve assuring results. They all prove that contrastive learning can bring more structural embedding space for pixels in different classes. But they also can be improved in several points. First, pixel-wise positive samples are easier to get compared to image-wise ones and can be fully utilized by proper loss function design. Second, anchor points are not grouped clearly. It is distinct that hard pixels have dissimilar distribution in embedding space with different score maps and ground-truth. The hard anchor should be regrouped by label and prediction.

\subsection{Distribution Optimization}
\label{2.3}

The strong relationship between pixels is an important property in image processing. Traditional pixel-wise cross entropy loss formulates it as an independent pixel labeling problem and ignores structure information between pixels. As shown in Fig.\ref{Figure.2}, the clearer distribution of features is, the higher accuracy of the model is. For the optimization of feature distribution, it can be roughly divided into two directions: one is to find a more accurate measurement method. Several ways to exploit the distribution information between pixels have been investigated \cite{ke2018adaptive,zhao2019region,jin2021mining, li2022deep, zhou2022rethinking}. RMI \cite{zhao2019region} models spatial structure in the 3x3 window and emphasizes adjacent neighborhoods. AAF \cite{ke2018adaptive} makes a breakthrough at distance computation between pixels than RMI. MCIBI \cite{jin2021mining} gathers pixel level feature distribution by class level memory bank which is updated during training step. 
\cite{li2022deep} proposed hierarchical semantic segmentation (HSS) to achieve a structured and pixel-wise description of visual observation in terms of a class hierarchy. In \cite{zhou2022rethinking}, a non-parametric prototype representation is used to replace the dense prediction task with the nearest prototype retrieval.
Another one is looking for more suitable teachers. A simple way is label smoothing \cite{szegedy2016rethinking} which improves the distribution of labels by regularization, and it reduces the difficulty of learning targets for model to prevent over-fitting. Furthermore, in some lightweight works, knowledge distillation \cite{hinton2015distilling} can bring well distribution of large model to lightweight model.

Combining the above two points, our work provides another perspective on distribution optimizing by deploying contrastive learning in supervised semantic segmentation. The method utilizes high-quality embedding of correctly classified pixels to optimize distribution of major pixels being classified falsely from proper class distribution. Furthermore, distribution optimizing is only used in the segmentation model during training step, without extra inference step during testing.

\section{Method}
\label{headings}
\subsection{Contrastive Learning}
\label{3.1}

{\bf{Supervised pixel-wise contrastive Learning}}. As common settings, an image $I \in \mathbb{R}^{3\times H\times W}$ is fed into an encoder-decoder architecture which exports feature maps $F \in \mathbb{R}^{C\times \tfrac{H}{4}\times \tfrac{W}{4}}$. Then, the feature maps will go through segmentation head and up-samplings to produce score map $S \in \mathbb{R}^{1\times H\times W}$ of the prediction. Last year, several works convert contrastive loss from image-size to pixel-size \cite{wang2021exploring,hu2021region}, and optimize the contrastive pairs sampling strategy. Positive pairs containing anchor $i \in \mathbb{R}^D$ and positives $i^{+}$ are the same class in ground-truth map $G \in \mathbb{R}^{1\times H\times W}$, but negatives $i^{-}$ are sampled from different classes in identical image or others. Generally, the conventional loss is defined as:
\begin{equation}
    \label{sup-loss}
    L^{NCE}_{i} =\cfrac{1}{|P_i|} \sum_{i^+\in {P_i}} -\mathtt{log} \cfrac{\mathtt{exp}(i\cdot i^{+}/\tau)}{\mathtt{exp}(i\cdot i^{+}/\tau)+\sum_{N_i} \mathtt{exp}(i\cdot i^{-}/\tau)},
\end{equation}
where for pixel embedding $i$, $P_i$ denotes the positive embedding set, $N_i$ denotes the negative embedding set and $|P_i|$ denotes numbers of positive pairs. ‘$\cdot$’ means the inner (dot) product.

{\bf{Project Head}}. SimCLR \cite{chen2020simple} adds MLP as the project head in contrastive learning for the first time. In subsequent works, MLP has been widely used because of its superiority. \cite{wang2021revisiting} does extensive research on MLP and verifies the effect of MLP on the embedding space on computer vision task.

{\bf{Loss Combining}}. As stated in Eq. \ref{sup-loss}, $L^{NCE}$ is designed to adjust the embedding space, and not to force the network to output symbolic results. Fortunately, the cross entropy loss used widely in semantic segmentation task can represent the difference between the sample label and the predicted probability. Following \cite{wang2021exploring}, the function can be formed as:
\begin{equation}
    \label{ce-loss}
    L^{CE}_{i} = -1^{\top}_\mathtt{\overline{c}}\mathtt{log(softmax}(y)),
\end{equation}

where $1_{\overline{c}}$ denotes the one-hot encoding of $\overline{c}$, the logarithm is defined as element-wise, and 

\begin{equation}
    \label{softmax}
    \mathtt{softmax}(y_c)={\mathtt{exp}(y_c)}/{\sum^{|C|}_{c^{\prime}=1}\mathtt{exp}(y_{c^{\prime}})}.
\end{equation}

\begin{figure*}[!t]
	\centering
	\label{Fig.2}
	\includegraphics[width=1\textwidth]{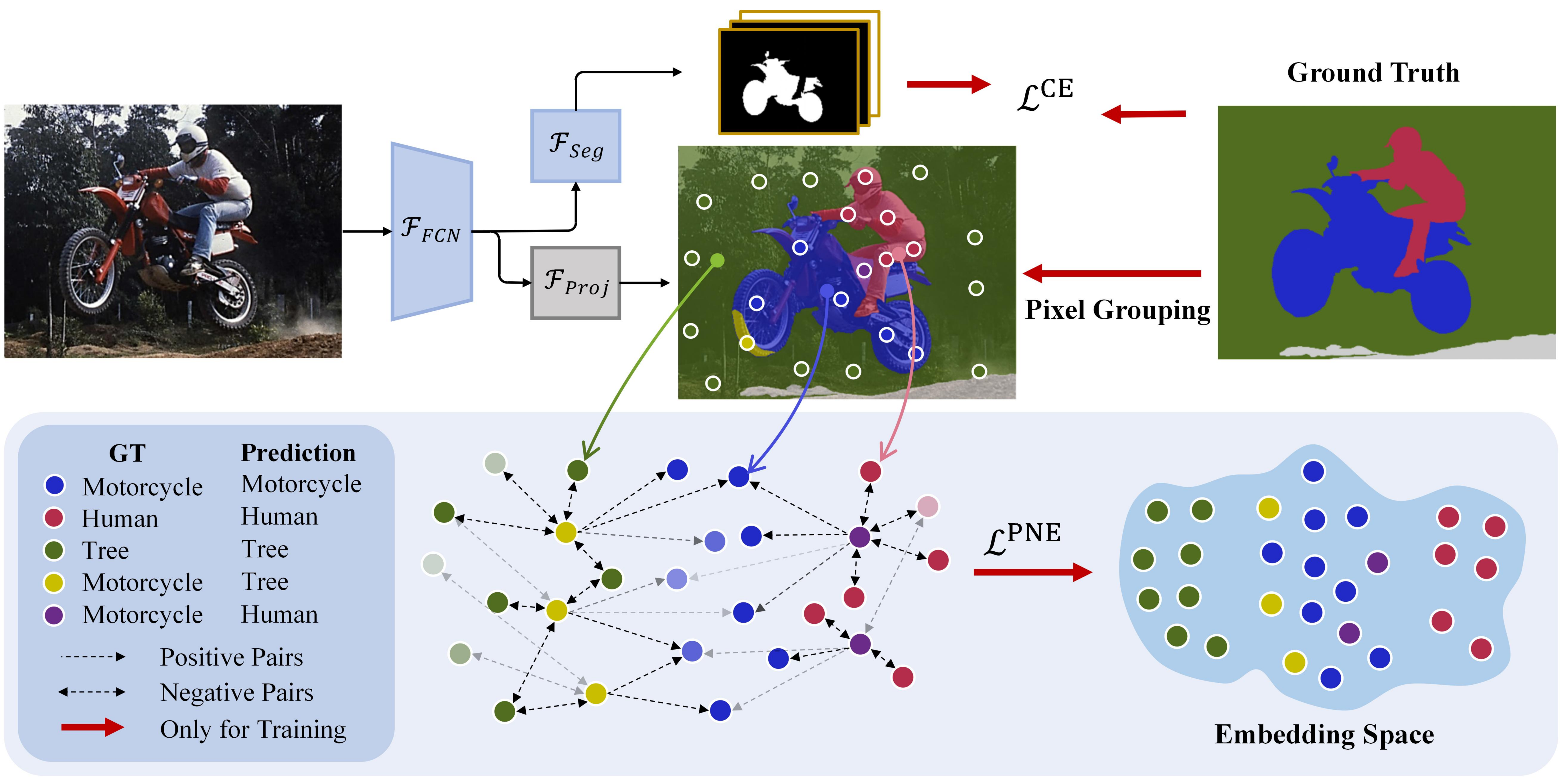}
	\caption{The overall architecture and design details of the proposed PNE loss.
	Pixel embedding extracted by $\mathcal{F}_{proj}$ and $\mathcal{F}_{FCN}$.
	For per class, pixel embeddings are grouped into correctly classified and each misclassified category according to Ground Truth (GT) and predictions from $\mathcal{F}_{seg}$. 
	The property of the misclassified pixel embeddings point out the classes of positive and negative samples. The anchors, positives and negatives are all randomly sampled from limited regions respectively.}
	\label{pic_arcitecture_print}
\end{figure*}

The supervisory information (from $G \in \mathbb{R}^{1\times H\times W}$) utilized by cross entropy loss is poor but distinct. Meanwhile, the information (from $F \in \mathbb{R}^{C\times \tfrac{H}{4}\times \tfrac{W}{4}}$) utilized by $L^{NCE}$ is obscure but abundant. It is obvious that their information types are complementary. As shown in Fig.\ref{pic_arcitecture_print}, cross entropy loss can provide simple but clear pixel categories, and $L^{NCE}$ can optimize the feature representation of anchor in high-dimension space and adjust the structural similarity between pixel embeddings. The combined function used for training the model is: 
\begin{equation}
    \label{combine-loss}
     L^{COM}_{i} = L^{CE}_{i}+\alpha L^{NCE}_{i},
\end{equation}
where the $L^{COM}_{i}$ can directly act on the segment head, and $\alpha$ is weight factor.

\subsection{Balance of Positives and Negatives}
{\bf{More attention to positives.}} As stated in \ref{2.3} and \ref{3.1}, supervised contrastive learning is limited by unsupervised paradigm that struggles with the lack of positive samples. However, with the help of labeled data, positive pairs sampled from the same class are not scarce anymore. \cite{wang2021understanding} proves that optimized gradient of $L_i^{NCE}$ like Eq.\ref{sup-loss} and Eq.\ref{ce-loss} is mostly from negative pairs. That situation will make model prefer to separate anchors from negatives rather than gather anchors and positives. 
Let us consider a normal situation for Eq.\ref{sup-loss}: to simplify, we refer $S_{i, i^{+},\tau}$, $S_{i, i^{-}, \tau}$ to $i\cdot i^{+}/\tau$, $i\cdot i^{-}/\tau$:
\begin{equation}
    \label{prove-equation}
    \begin{aligned} 
    L^{NCE}_{i} &=\cfrac{1}{|P_i|} \sum_{i^{+}\in {P_i}} -\mathtt{log} \cfrac{\mathtt{exp}(S_{i, i^{+},\tau})}{\mathtt{exp}(S_{i, i^{+},\tau})+\sum_{i^{-}\in{N_i}} \mathtt{exp}(S_{i, i^{-},\tau})}\\
                &=\cfrac{1}{|P_i|} \sum_{i^+\in {P_i}}\mathtt{log}( \cfrac{\mathtt{exp}(S_{i, i^{+},\tau}) +|N_i| \ \mathbb{E}_{i^-\in{N_i}}[\mathtt{exp}(S_{i, i^{-},\tau})]}{\mathtt{exp}(\mathbb{E}_{i^+\in {P_i}}[S_{i, i^{+},\tau}])})\\
                &=\cfrac{1}{|P_i|} \sum_{i^+\in {P_i}}\mathtt{log}(1+ \cfrac{\mathtt{exp}(S_{i, i^{+},\tau}) - \mathtt{exp}(\mathbb{E}_{i^+\in {P_i}}[S_{i, i^{+},\tau}]) +|N_i| \ \mathbb{E}_{i^-\in{N_i}}[\mathtt{exp}(S_{i, i^{-},\tau})]}{\mathtt{exp}(\mathbb{E}_{i^+\in {P_i}}[S_{i, i^{+},\tau}])})\\
                &\approx \mathtt{log}(1 + \cfrac{|N_i|\mathbb{E}_{i^-\in{N_i}}[\mathtt{exp}(S_{i, i^{-},\tau})]}{\mathtt{exp}(\mathbb{E}_{i^+\in {P_i}}[S_{i, i^{+},\tau}])}), \qquad \mathtt{if} \ \tau \neq singular \ value \\
    \end{aligned}
\end{equation}
where $|P_i|$ and $|N_i|$ are numbers of positive and negative pairs. The value of $S_{i, i^{+}, \tau}$ and $S_{i, i^{-}, \tau}$ are both in $[-1/\tau, 1/\tau]$. With comparatively large $|N_i|$, it is obvious that negative pairs provide most of the penalty value.
The value of NCE loss will be decided by the number of negative pairs.

As shown in Fig.\ref{Figure.2}, the distance between classification centers and the separability of category boundaries are both important for classification task. Therefore, we need to improve the form of $L^{NCE}$ to cluster the embedding of every category. With this insight, we put more weight to positive pairs in logarithm function by transferring summation of positives. After simplified, our loss function is defined as:
\begin{equation}
    \label{change-loss}
    L^{PNE}_{i} =\mathtt{log}(1+ \cfrac{\sum_{i^-\in N_i}\mathtt{exp}(i\cdot i^-/\tau)}{\sum_{i^+\in P_i}\mathtt{exp}(i\cdot i^+/\tau)}).
\end{equation}

As Eq.\ref{prove-equation} setting, the Eq.\ref{change-loss} can be simplified to: 
\begin{equation}
    \label{prove-equation1}
    \begin{aligned} 
    L^{PNE}_{i} &=\mathtt{log}(1+ \cfrac{\sum_{n\in N_i}\mathtt{exp}(S_{i, i^{-},\tau})}{\sum_{p\in P_i}\mathtt{exp}(S_{i, i^{+},\tau})}) \\
                &=\mathtt{log}(1+\cfrac{|N_i| \ \mathbb{E}_{i^-\in{N_i}}[\mathtt{exp}(S_{i, i^{-},\tau})]}{|P_i| \ \mathbb{E}_{i^+\in{P_i}}[\mathtt{exp}(S_{i, i^{+},\tau})]})\\
                &=\mathtt{log}(1+\cfrac{\mathbb{E}_{i^-\in{N_i}}[\mathtt{exp}(S_{i, i^{-},\tau})]}{\mathbb{E}_{i^+\in{P_i}}[\mathtt{exp}(S_{i, i^{+},\tau})]}), \qquad \mathtt{if} \ |N_i|=|P_i| \\
    \end{aligned}
\end{equation}

Eq.\ref{prove-equation} has proved that the number of negative samples plays a domain role in the original contrastive loss. However, the contrastive loss should focus on the relationship between the mean distance from the anchor to the negative samples and the mean distance from the anchor to the positive samples.
Therefore, we set $\ |N_i|=|P_i|$ to treat negative and positive pairs equally. The improved PNE loss ignores the influence of pairs number and is only calculated by the cosine distance of pairs, which is intuitively more reasonable.

{\bf{Per-positive Weight.}}  
Following Eq.\ref{change-loss}, we construct a region-to-pixel contrastive learning strategy with positive and negative pairs.
In this way, all samples can be utilized for the optimization of the network. 
However, there is different importance in each sample.

For positives, we introduce prediction scores to adjust the weights of each positive sample since the confidence of different positive samples is completely different. Positive samples with high confidence may provide more accurate guidance for the anchor. Fortunately, the predicted score naturally reflects confidence in the correct prediction of the pixel. With this idea, we use a prediction score $w_{i^+}$ after $\ softmax$ to multiply with the cosine distance of positive pair $(i, i^+)$ as the similarity measure, which means that high-quality positive features will be paid more attention. 
The mathematical expression can be defined as:
\begin{equation}
\label{deqn_ex1a}
L^{PNE}_{i} =\mathtt{log}(1+ \cfrac{\sum_{i^-\in N_i}\mathtt{exp}(i\cdot i^-/\tau)}{\sum_{i^+\in P_i}w_{i^+}/\Bar{w}_+\cdot\mathtt{exp}(i\cdot i^+/\tau)}),
\end{equation}
where $\Bar{w}_+$ represent the average of $w_{i^+}$.

But for negatives, refinement of every negative pair is not necessary. 
Individually, the cosine similarities of negative pairs have already decided gradient contribution from every negative sample to specific anchor. 
The feature quality of single negative sample has no obvious significance of reference for anchor from other categories. As a whole, anchors need to be far away from all of them in embedding space. The weights change of single pairs do not contribute much to the entirety of negative samples. 
We provide the corresponding experiments of the positive weighting in ablation study.

\subsection{Construction Strategies}
\label{samplingstrategy}

\subsubsection{Individual Anchor Sets} 
Directly converting contrastive learning from image-wise to pixel-wise faces computation limits without sampling. Previous works provide several feasible ways which focus on improving the discrimination power of models by sampling hard anchors (the pixel with incorrect predictions). However, these strategies still exist embedding confusion and computational waste in negative pairs. As shown in Fig.\ref{pic_arcitecture_print}, hard anchors, containing different individual set $S_{l,k}$($l$ means predicted result, $k$ means ground-truth label), are classified as different classes $\mathbb{\widetilde{C}}$ ($l\in\mathbb{\widetilde{C}}$, and $ \mathbb{\widetilde{C}} \subset \mathbb{C}\backslash k$, where $\mathbb{C}\backslash k$ denotes the pixels belonging to the classes except class $k$).
They are distributed in different embeddings space, even though have same class $k$ in ground-truth. It is unreasonable that the hard anchors in different distribution are guided by same negative samples.
In our method, hard anchors with same ground-truth label but different predicted results are divided into individual sets which will be calculated separately, not mixed up anymore.
    
\subsubsection{Corresponding Negatives} 
In semantic contrastive learning \cite{wang2021understanding,robinson2020contrastive}, negative samples should select pixels similar to the anchor as much as possible. During calculating contrastive loss, utilizing difficult-to-distinguish negative samples can optimize pixel embedding space faster.
In this paper, we use semantically explicit category information to select difficult negative samples instead of cosine similarity used in previous works \cite{wang2021understanding,robinson2020contrastive}. 
Cosine distance can not accurately measure embedding similarity which usually refers to surrounding structural information and the module of a vector in addition. Instead of using cosine similarity, it is better to select difficult negative samples by obvious classificatory information from label. 
In details, for samples with the same predictions, their similarity is naturally closer than others. 
The individual anchor set $S_{l,k}$ are more similar to $i_{l,l}$ (pixel set that left $l$ means predicted result and right $l$ means label) which have same predicted result. 
Thence, corresponding negative set $N_{l,l}$ for individual anchor set $S_{l,k}$ is only sampled from $i_{l,l}$ in PNE loss.

In this way, the negative pairs, constructed by these individual anchor sets and corresponding negatives, will be calculated more simply without filtering anchor and negatives by their difficulty. Furthermore, the anchors will be optimized more pertinently by their corresponding negatives. Meanwhile, all of anchors will not lose context information contrast to other contrastive loss because all relevant negatives were calculated with a pair scale. Thus, the supervised contrastive loss is defined as:
\begin{equation}
\label{deqn_ex1a}
L^{PNE}=\cfrac{1}{|S|} \sum_{l\in C}\sum_{k\in C\backslash l}\sum_{ i\in S_{l,k}} \mathtt{log}(1+ \cfrac{\sum_{i^-\in N_{l,l}}\mathtt{exp}(i\cdot i^-/\tau)}{\sum_{i^+\in P_{k,k}}w_{i^+}/\Bar{w}_+\cdot\mathtt{exp}(i\cdot i^+/\tau)}),
\end{equation}
where $|S|$ denotes the numbers of pixels with incorrect predictions. Similar as $N_{l,l}$, $P_{k,k}$ are sampled from $i_{k,k}$. $|S|<200$ is set to save computation resources.

\section{Experiment}
\label{others}
In this section, we conduct comprehensive experiments to evaluate the PNE loss on three popular segmentation datasets, including Cityscapes, COCO-Stuff, and ADE20K. Experiments demonstrate that the PNE loss achieves competitive performances. In the following parts, we first introduce the experimental datasets, evaluation metrics, and implementation details, and then compare the best results with state-of-the-art methods on Cityscapes, ADE20K, and COCO-Stuff. At last, we will present a series of ablation experiments performed on the Cityscapes dataset.  

\subsection{Experimental Datasets}
        \textbf{Cityscapes} is a large-scale dataset for segmentation task, which is collected from 50 different urban street scenes. The dataset contains 5K finely annotated images and 20K coarsely annotated images of driving scenes. 
        The 5K finely annotated images used in our experiment are divided into 2,975, 500, and  1,525 images for training, validation, and testing respectively. The label of this dataset defines 19 categories of different objects for the dense pixel prediction task, reaching 97\% coverage per image. Especially, We only use the train set (2,975 images) for training and both the val and test set for evaluation.
        
        \textbf{ADE20K} is a widely-used semantic segmentation dataset, containing up to 25K images in 150 categories that are densely annotated. The 25K images are split into sets with numbers 20K, 2K, and 3K for training, validation, and testing. The image scene is diverse and contains various categories of objects, which makes the dataset more challenging. 
        
        \textbf{COCO-Stuff} is a large-scale benchmark that is applied for instance segmentation as well as semantic segmentation, providing rich annotations for 91 thing classes and 91 stuff classes. The dataset has 10K images, which are partitioned into 9K images for training and 1K images for validation. 

\subsection{Evaluation metric and Implementation Details}
\subsubsection{Evaluation metric} 
The standard mean Intersection of Union (mIoU) is chosen as the evaluation metric. Following general protocol \cite{wang2021exploring,hu2021region}, we average the segmentation results over multiple scales with flipping, i.e., the scaling factor is 0.75 to 2.0 (with the interval of 0.25). For ablation studies, the single-scale evaluation is performed if not mentioned.

\subsubsection{Implementation Details}
We select DeepLabV3, OCRNet, HRNetV2, and UperNet as our baseline, and apply HRNetV2, ResNet-101, and  Swin-B as backbone pre-trained on ImageNet. PNE loss is developed on mmsegmentation toolbox \cite{mmseg2020}, and attached to DeepLabV3, OCRNet, and Swin-Transfomer to validate the generalization ability. NVIDIA GeForce GTX 3090 (24GB memory) is used in our experiments for training models. Additionally, we plug the PNE loss into the position behind the bottleneck as stated in Sec. \ref{samplingstrategy} and implement our method based on Pytorch \cite{paszke2019pytorch}. Following \cite{wang2021exploring,hu2021region}, our model is trained with Stochastic Gradient Descent (SGD). During the training phase, a poly learning rate policy is employed where the initial learning rate is multiplied by (1-iter/total\_iter) 0.9 after each iteration. The base learning rate is set to 0.001 for COCO-Stuff and 0.01 for Cityscapes and ADE20K. Momentum and weight decay coefficients are set to 0.9 and 0.0005 respectively. The batch size is set to 8 for Cityscapes and 16 for COCO-Stuff and ADE20K. 
Meanwhile, the training period is set to 40K steps for Cityscapes, 60K steps for COCO-Stuff, and 160k steps for ADE20K. 
In addition, random scaling, random cropping, and random left-right flipping are applied as data augmentation during training.

\begin{table}[htbp]
\caption{The details of experimental settings for the three benchmark datasets.}
\scriptsize
    \begin{center}
    \label{experiments_config}
    \begin{tabular}{c|c|c|c}
    \toprule 
    Config             & Cityscapes   & COCO-Stuff & ADE20K \\\hline  
    training iterations    & 40,000     & 60,000           & 160,000    \\
    learning rate      & 0.01      & 0.001        & 0.01   \\
    batch size         & 8             & 16               & 16    \\
    weight decay rate  & 0.0005    & 0.0005        & 0.0005    \\
    \bottomrule 
    \end{tabular}
    \end{center}
\end{table}

\subsection{Comparison to State-of-the-Arts}

\subsubsection{Quantitative and Qualitative Analysis on Cityscapes}
\begin{table}[htpb]
\caption{Results on the Cityscapes val and test set.}
\scriptsize
    \begin{center}
    \begin{tabular}{r |c |c |c |c }
    \toprule[1.5pt]
    Method & Backbone & sec./iter. & val mIoU$\%$ & test mIoU$\%$ \\
    \hline
    AAF\cite{ke2018adaptive} & D-ResNet-101    & - & 79.2 & 79.1 \\ DeepLabV3+\cite{chen2018encoder} & D-Xception-71 & -& 79.6 & -\\
    PSPNet\cite{zhao2017pyramid} & D-ResNet-101  & -& 79.7 & 78.4 \\ 
    DANet\cite{fu2019dual} & D-ResNet-101 & - & 81.5 & -\\
    HANet\cite{choi2020cars} & D-ResNet-101    & -& 80.3 & - \\
    SpyGR\cite{li2020spatial}  & D-ResNet-101       & -& 80.5 & -\\
    ACF\cite{zhang2019acfnet} & D-ResNet-101         & -& 81.5 & -\\
    \hline
    EncNet\dag\cite{zhang2018context} & D-ResNet-101 & - & 77.1 & - \\
    EncNet + Ours & D-ResNet-101 & - & \textbf{77.7(+0.6)} & - \\
    \hline
    DeepLabV3\cite{chen2017rethinking} & D-ResNet-101 & 1.92 & 78.5 & 78.1 \\
    DeepLabV3 + Ours & D-ResNet-101 & 2.43 & \textbf{80.8(+2.3)} & \textbf{79.3(+1.2)}\\ 
    \hline
    HRNetV2\cite{sun2019high} & HRNetV2-W48 & 0.57 & 79.7 & 79.4\\
    HRNetV2 + Ours & HRNetV2-W48 & 0.64 & \textbf{81.5(+1.8)} & \textbf{80.6(+1.2)}\\
    \hline
    OCRNet\cite{yuan2020object} & HRNetV2-W48 & 1.40 & 81.6 & 80.4\\
    OCRNet + Ours & HRNetV2-W48 & 1.65 & \textbf{82.9(+1.3)} & \textbf{81.7(+1.3)}\\
    \bottomrule[1.5pt]
    \end{tabular}
    \end{center}
\label{city-val}
\end{table}

We present the evaluation results of the proposed method on Cityscapes validation and test set and compare them with other state-of-the-art methods in Table \ref{city-val}. We find that, with a small increase in training time, the accuracy of model on the validation and test set can be improved considerably without sacrificing inference time. Specifically, the performance of DeepLabV3 improved $\textbf{2.3\%}$ mIoU on validation set from 78.5$\%$ to 80.8$\%$ and $\textbf{1.2\%}$ mIoU on test set from 78.1$\%$ to 79.3$\%$ by PNE loss.
Simultaneously, OCRNet with our method yields the mIoU of 82.9$\%$ and 81.7$\%$ on validation and test set respectively. 
Besides, EncNet sets up an auxiliary classification task to make full use of the prior knowledge of the scene and reuse the classification logits to highlight the category information associated with the scene. This may have some similarities and conflicts with contrastive learning which optimizes the distribution of pixel embeddings of different categories in feature space. However, PNE loss can provide more fine-grained category information guidance at the pixel level, making up for the shortcomings of EncNet and achieving non-negligible performance improvements (0.6\% mIoU).
Both of the improvements demonstrate the proposed PNE loss can effectively promote the performance of the baseline as expected.
To be mentioned, the settings of all the above testing results are based on multi-scale and flipping tricks.

\subsubsection{Quantitative and Qualitative Analysis on COCO-Stuff}

\begin{table}[htpb]
\caption{Results on the COCO-Stuff test set.}  
\label{coco}
\scriptsize
    \begin{center}
    \begin{tabular}{r |c |c }
        \toprule[1.5pt]
        Method & Backbone & mIoU$\%$ \\
        \hline
        SVCNet\cite{ding2019semantic} & D-ResNet-101 & 39.6 \\
        DANet\cite{fu2019dual} & D-ResNet-101 & 39.7 \\
        SpyGR\cite{li2020spatial} & ResNet-101 & 39.9 \\
        ACNet\cite{fu2019adaptive} & ResNet-101 & 40.1 \\
        \hline
        DeepLabV3\dag\cite{chen2017rethinking} & D-ResNet-101  & 38.8 \\
        DeepLabV3 + Ours & D-ResNet-101  & \textbf{39.5(+0.7)} \\
        \hline
        OCRNet\cite{yuan2020object} & 
        HRNetV2-W48 & 40.5 \\  
        OCRNet + CIPC\cite{wang2021exploring} & HRNetV2-W48 & 41.0(+0.5)\\      
        OCRNet + Ours & HRNetV2-W48 & \textbf{41.2(+0.7)}\\
        \hline
        UperNet\cite{xiao2018unified} & Swin-B\cite{liu2021swin} & 43.7 \\
        UperNet + Ours & Swin-B & \textbf{44.3(+0.6)}            \\
        \bottomrule[1.5pt]
    \end{tabular}
    \end{center}
\end{table}

\begin{figure}[htpb]
    \centering
    % \flushleft
    \includegraphics[width=1\textwidth]{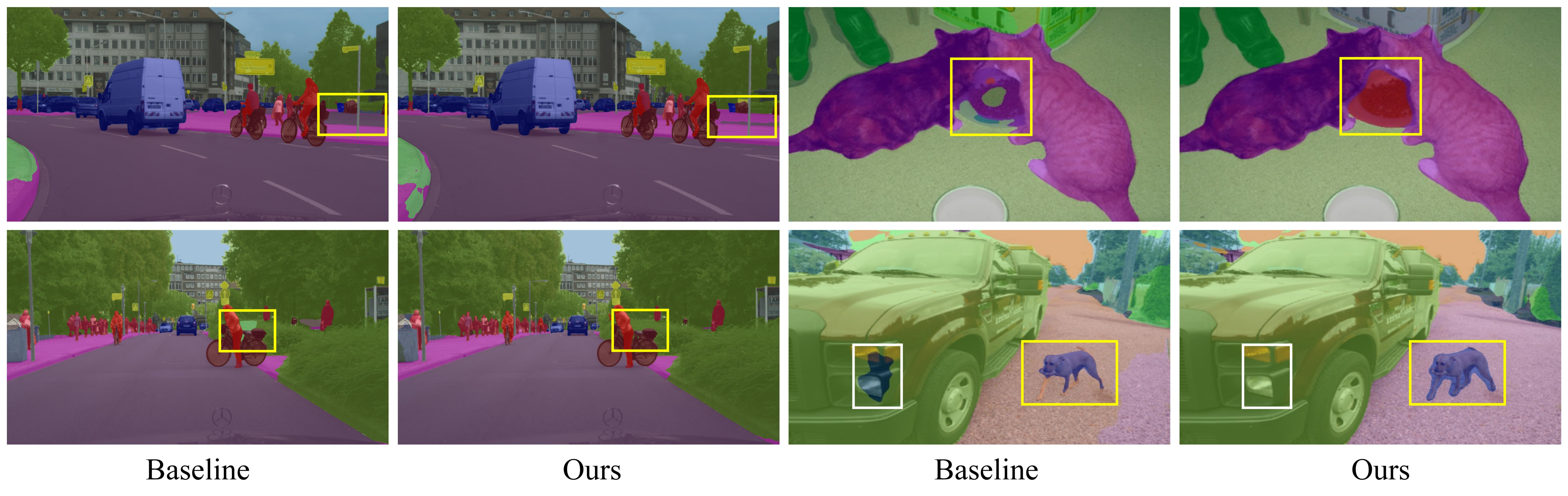}
    \caption{Visual comparison between baseline and ours (left: Cityscapes, right: COCO-Stuff).}
    \label{Figure.3}
\end{figure}

We also conduct experiments on COCO-Stuff to verify the effectiveness of PNE loss.
Table \ref{coco}, where \dag \ denotes results posted by the mmsegmentation toolbox \cite{mmseg2020}, presents excellent results produced by PNE loss on the COCO-Stuff test set.
Our method outperforms DeeplabV3 by $\textbf{0.7\%}$ mIoU, OCRNet by $\textbf{0.7\%}$ mIoU and Swin-B by $\textbf{0.6\%}$ mIoU. The great improvements illustrate that PNE loss can guide the baseline more effectively to capture complicated contextual representation. 
Compared with the previous contrastive method \cite{wang2021exploring}, our method surpasses $\textbf{0.2\%}$ mIoU. 
The improvement shows that PNE loss is more suitable for segmentation tasks.
It is worth noting that the reason why our suppose performance has not been greatly improved is that CIPC \cite{wang2021exploring} employs the memory bank to establish the context of the training data.
Since the memory bank in MoCo \cite{he2020momentum,chen2020improved} and CIPC \cite{wang2021exploring} needs to be implemented by momentum update, this will corrupt the plug-and-play characteristic of PNE loss. 
Considering the characteristics of loss function compatibility, our method of abandoning the memory bank makes a trade-off between accuracy and convenience.
However, PNE loss can also utilize the memory bank to preserve intra/inter-image pixel embeddings and foreseeably reach further performance.
In addition, PNE loss can not only boost the performance of network based on the CNN method, but also works on the model based on Transformer. 
Besides, the visual comparison also proves the improvement with PNE loss. As shown in Fig.\ref{Figure.3}, our method achieves more precise prediction than the baseline (like the region of the dog in the right second row).

\subsubsection{Quantitative and Qualitative Analysis on ADE20K}
\begin{table}[htpb]
\caption{Results on the ADE20K val set.}  
\label{ADE20K}
\scriptsize
    \begin{center}
    \begin{tabular}{r |c |c }
        \toprule[1.5pt]
        Method & Backbone & mIoU$\%$ \\
        \hline
        CFNet\cite{zhang2019co} & D-ResNet-101 & 44.89 \\
        CCNet\cite{huang2019ccnet} & D-ResNet-101 & 45.22 \\
        GANet\cite{zhang2019deep} & D-ResNet-101 & 45.36 \\
        \hline
        OCRNet\cite{yuan2020object} & HRNetV2-W48 & 45.66 \\
        OCRNet + CIPC\cite{wang2021exploring}& HRNetV2-W48 & 46.41(+0.75)\\  
        OCRNet + Ours & HRNetV2-W48 & \textbf{46.76(+1.10)} \\
        \bottomrule[1.5pt]  
    \end{tabular}
    \end{center}
 \end{table}
 
In our experiments, dataset ADE20K is used to prove the ability of PNE loss on large-scale images dataset. As expected, models trained with PNE loss show better-capturing ability of context in scene parsing and reach superior results compared to those without PNE loss. As shown in Table \ref{ADE20K}, the improvements of our method are $\textbf{1.10\%}$ mIoU in OCRNet.

\subsection{Visual Comparison}
\subsubsection{Visual comparison of pixel embeddings by t-SNE}

\begin{figure}[htpb]
    \centering
    \includegraphics[width=0.45\textwidth]{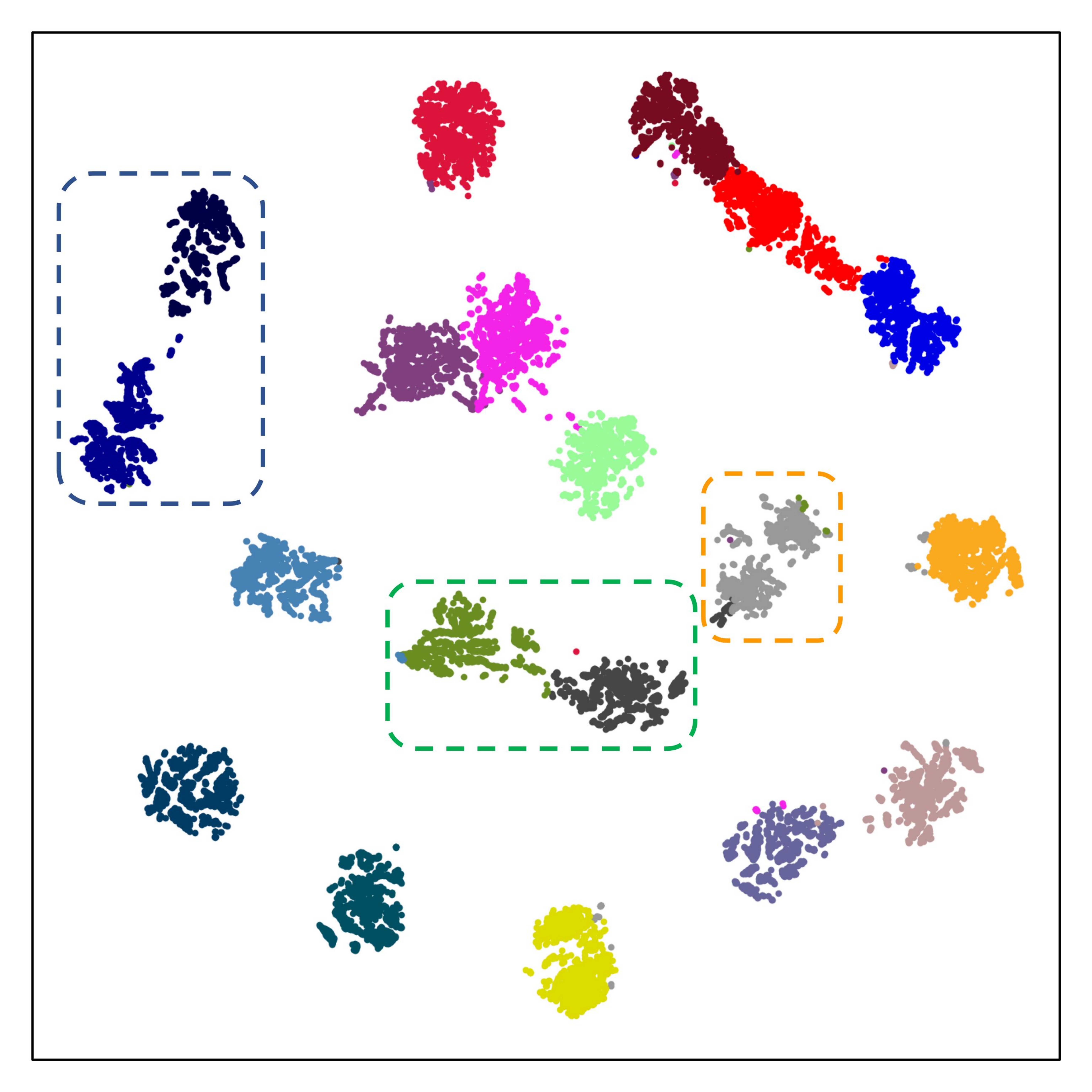}
    \caption{Cross entropy loss.}
    \label{fig_stne_baseline}
\end{figure}

\begin{figure}[htpb]
    \centering
    % \flushleft
    \includegraphics[width=0.45\textwidth]{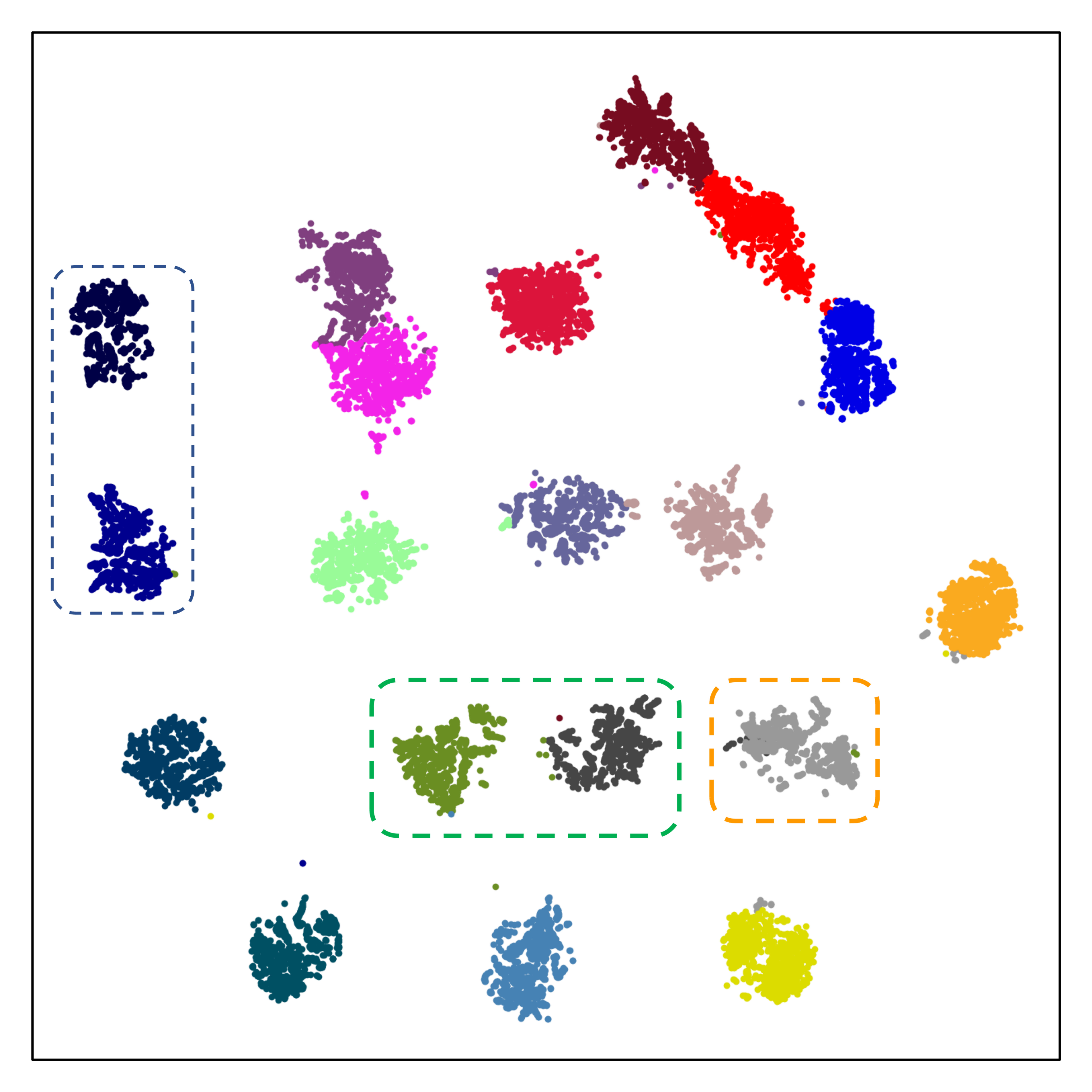}
    \caption{PNE loss and cross entropy loss.}
    \label{fig_stne_PNE}
\end{figure}

To prove the validity of PNE loss proposed, we visualize the distribution of pixel embeddings, as shown in Fig.\ref{fig_stne_baseline}. Each point represents the feature representation of a pixel. The similarity of pixel is exhibited by the distance between each point.
The uniformity(inter-class) and alignment(intra-class) of pixel embeddings are obtained substantial advance after plugging the PNE loss.
For uniformity, {\color[rgb]{0.4196,0.5569,0.1373}$\bullet$} (vegetation) and {\color[rgb]{0.2745,0.2745,0.2745}$\bullet$} (building) are separated more clearly with our method.
For alignment, {\color{gray}$\bullet$} (pole) are divided into three parts in different context, but aggregated unitedly as a whole in the Fig.\ref{fig_stne_PNE} with more closeness. That greatly demonstrates PNE loss can assist model to capture the relationships between pixels and improve segmentation accuracy.

\subsection{More visual comparison of semantic segmentation}
\label{headings}
\begin{figure}[htpb]
    \centering
    \includegraphics[width=0.9\textwidth]{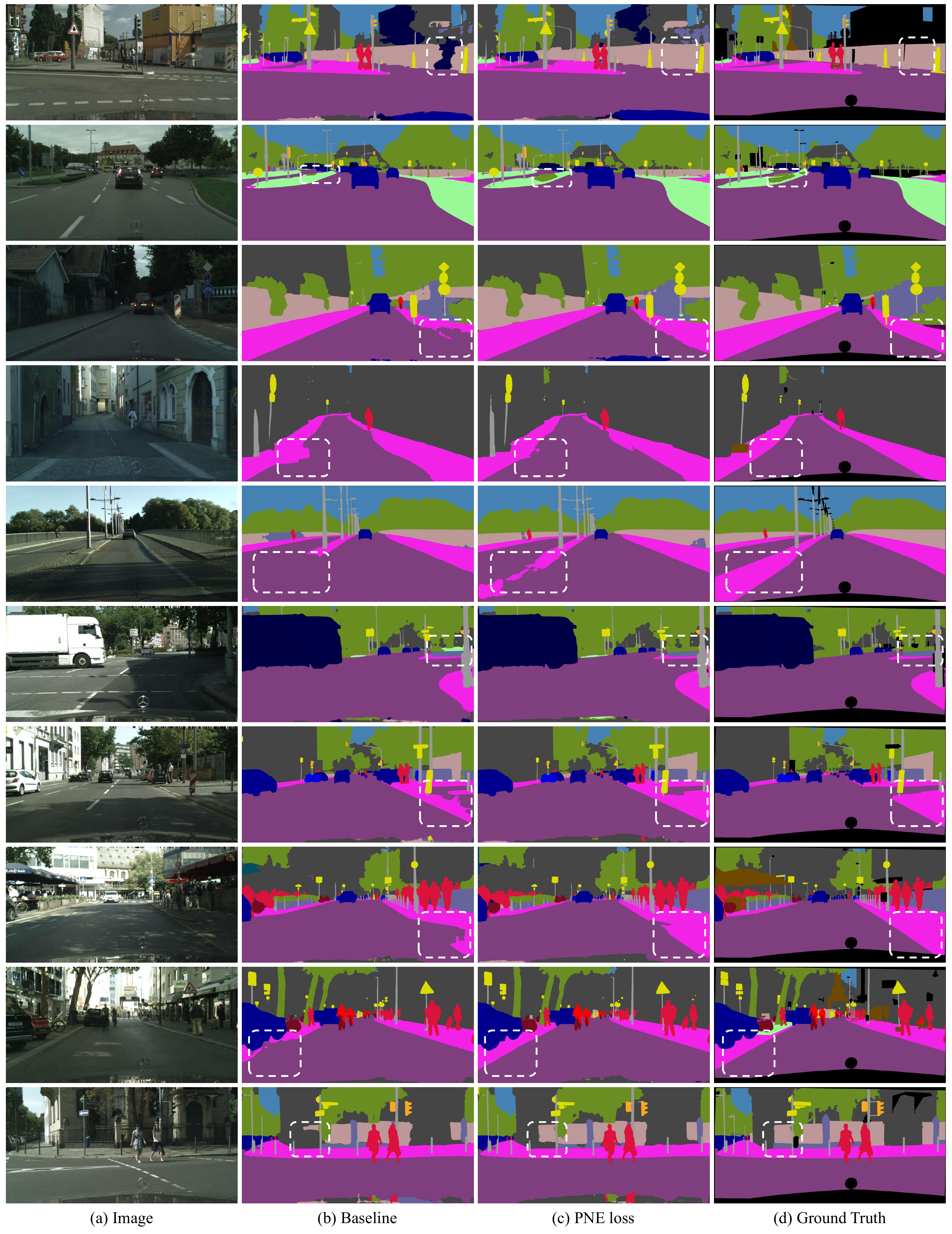}
    \caption{More visual comparison of segmentation results on Cityscapes.}
    \label{Cityscapes_visual_comparison}
\end{figure}

\begin{figure}[htpb]
    \centering
    % \flushleft
    \includegraphics[width=0.9\textwidth]{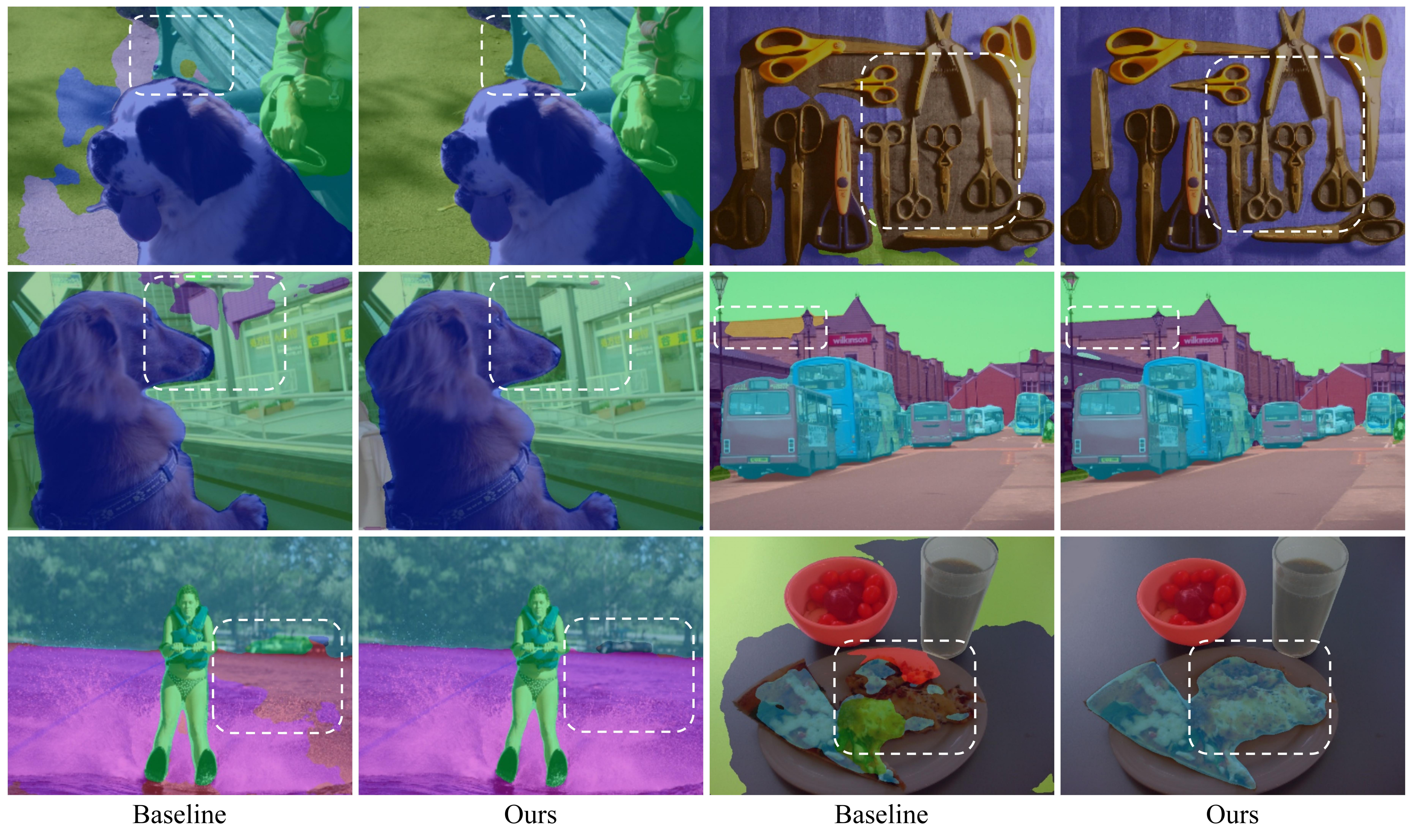}
    \caption{More visual comparison of segmentation results on COCO-Stuff.}
    \label{More_visual_comparison}
\end{figure}

To qualitatively show the advantages of our method, we provide a lot of visual comparisons which consist of prediction results in Cityscapes and COCO-Stuff. As shown in fig.\ref{More_visual_comparison} and fig.\ref{Cityscapes_visual_comparison}, our method obtains precise segmentation and results closer to ground truth.

\subsection{Ablation Study}
{\label{Ablation Study}}
In this subsection, extensive ablation studies are performed on the Cityscapes validation set to verify the effectiveness of our ideas. 
Unless explicitly specified otherwise, all experiments are conducted on the DeepLabV3 with  Dilated-ResNet-50 backbone. We choose DeepLabV3-R50 with cross-entropy loss function as our baseline.

\subsubsection{Ablation Study for Equal calculation} 
\begin{table}[htpb]
    \scriptsize
    \caption{Ablation study for effectiveness of contrast. D-ResNet denotes the dilated-ResNet with the output stride = 8.}
    \label{table1}
    \begin{center}
    \begin{tabular}{c | c |c}
        \toprule[1.5pt]
        Pixel Contrast      & Backbone      & mIoU$\%$ \\
        \hline
        Base (w/o Contrast)  & D-ResNet-50   & 76.4 \\
        Base (w/o Contrast)  & D-ResNet-101   & 78.5 \\
        \hline
        Asymmetric Contrast (Eq.\ref{sup-loss}) & D-ResNet-50    &  77.5(+1.1) \\
        Asymmetric Contrast (Eq.\ref{sup-loss}) & D-ResNet-101    &  79.1(+0.6) \\
        PNE Contrast  & D-ResNet-50    &  \textbf{80.3(+3.9)} \\
        PNE Contrast  & D-ResNet-101    &  \textbf{80.8(+2.3)} \\
        \bottomrule[1.5pt]
    \end{tabular}
    \end{center}
 \end{table}

We carry out experiments to evaluate the effectiveness of the proposed positive-negative equal contrast. As shown in Table \ref{table1}, the model with PNE contrastive loss brings the improvement of $\textbf{3.9\%}$ mIoU than baseline. No supervised, traditional contrastive loss in Eq.\ref{sup-loss} can also make mIoU rise from 76.4\% to 77.5\% by optimizing the distribution of feature space during training.  
However, compared with traditional contrastive loss, PNE contrastive loss greatly helps obtain significant improvement at least $\textbf{2.8\%}$ mIoU.
Further, experiments utilizing a large backbone (D-ResNet-101) are explored and gather similar results to baseline. Traditional contrastive loss and PNE contrastive loss improve mIoU by 0.6\% and 2.3\% respectively.
That all illustrates positive-negative equal contrast is more suitable than normal contrastive loss, which is widely applied on unsupervised way in semantic segmentation.

\subsubsection{Ablation Study for Construct Strategy} 
\begin{table}[htpb]
    \scriptsize
    \caption{Ablation study for construct strategy of negative pairs. IAS and CN represent individual anchor sampling and corresponding negative.}
    \label{table2}
    \begin{center}
    \begin{tabular}{c c c |c}
        \toprule[1.5pt]
        Construct Strategy   & IAS   & CN    & mIoU$\%$ \\
        \hline
        PNE Contrast         &     &     & 78.8 \\
        Asymmetric Contrast        &     &     & 77.5 \\
        \hline
        PNE Contrast         & $\surd$ & &  79.3(+0.5) \\
        PNE Contrast         &  & $\surd$ &  79.6(+0.8) \\
        PNE Contrast         & $\surd$ & $\surd$ &  \textbf{80.3(+1.5)} \\
        Asymmetric Contrast         & $\surd$ & $\surd$ &  79.0(+1.5) \\
        \bottomrule[1.5pt]
    \end{tabular}
    \end{center}
 \end{table}

Here we validate the construct strategy of negative pairs and anchors. 
The results are summarized in Table \ref{table2}. As expected, individual anchor sets (IAS) and corresponding negatives (CN) all improve the performance of baseline, but it makes more sensitive to combine them to construct negative pairs. 
In details, the improvement of 0.5\% mIoU can be obtained by the individual anchor sets.
The main reason is that the model optimization direction comes from individual anchors according to class information instead of mixed anchors that contain predictions of different categories. 
The feature embedding distribution of mixed anchors is scattered. 
Constructing a contrastive loss for individual anchors is beneficial for model learning. Furthermore, the construct strategy of corresponding negatives also makes performance increase 0.8\% mIoU due to more explicit negative pairs compared to sampling based on cosine similarity. Combining the two strategies achieves a performance improvement of 1.5\% mIoU. IAS separates the mixed anchor and CN provides the corresponding negative samples accurately. The interaction between IAS and CN yields great effect.
For a fair comparison, we replace the baseline of cross entropy loss with asymmetric contrast and PNE contrast. 
Moreover, the results demonstrate that our construct strategy of negative pairs is suitable for PNE contrast.

\subsubsection{Ablation Study for Per-positive weights} 
\begin{table}[htpb]
\scriptsize
    \caption{Ablation study for Per-positive weights. PW Represents Per-positive weights.}
    \label{table3}
    \begin{center}
      \begin{tabular}{c c |c}
        \toprule[1.5pt]
        Per-positives weights   & standard deviation  & mIoU$\%$ \\
        \hline
        PNE Contrast (w/o PW ) & - & 80.1 \\
        \hline
        PNE Contrast ( PW w/o $\ softmax$) & 1.547 &  80.1 \\
        PNE Contrast ( PW w $\ softmax$)  & 0.112 & \textbf{80.3}\\
        \bottomrule[1.5pt]
    \end{tabular}     
    \end{center}

 \end{table}

We also implement experiments to evaluate the significance of Per-positive weights. And then, we measure the influence on probability score map from softmax. As shown in Table \ref{table3}, our method reaches better results with per-positive weights. Proving that the measure of similarity is not enough to fully utilize positive samples. Positive weights make the anchor pay more attention to the features with better separability. Similar to poor students learning directly from top students to get higher grades. Furthermore, the probability is the direct output of the network and the standard deviation is large, which is not suitable  to weight normalization.  
Fortunately, $\ softmax$ can resolve numerical problems while generating more stable weights and maintaining discrimination.

\subsubsection{Influence of hyper-parameters $\tau$ and $\alpha$} 

\begin{table}[htpb]
\scriptsize
    \caption{Ablation study for temperature $\tau$ and PNE contrastive loss weight $\alpha$.}
    \label{table4}
    \begin{center}
    \begin{tabular}{c |c c c c c c c c c c}
        \toprule[1.5pt]
        $\tau$      & 0.1  & 0.3 & 0.6  & 0.9 & 1 & 2 & 5 & 10\\
        mIoU$\%$   & 4.2 & 79.3 & 79.7 & 80.1 & \textbf{80.3} & 80.0 & 79.4 & 79.3 \\
        \hline
        $\alpha$      & 0.4  & 0.6 & 1.2 & 1.3 & 1.5 & 1.9 & 2.0 & 2.3 \\
        mIoU$\%$   & 79.2 & 79.4 & 79.7 & \textbf{80.3} & 79.7 & 79.2 & 79.3 & 79.4  \\
        \bottomrule[1.5pt]
        \end{tabular}
    \end{center}
 \end{table}

we carry out experiments to assess the influence of hyper-parameters (temperature $\tau$ and PNE contrastive loss weight $\alpha$). 1) Temperature $\tau$ has been investigated thoroughly and it is a consensus that $\tau$ controls the attention from loss to hard example and affects performance easily. 
However, in Table \ref{table4}, it shows that our loss is not sensitive to temperature $\tau$. We analyze that PNE contrast has filtered hard positive and negative pairs before calculating with temperature $\tau$. 2) It can be clearly seen that $\alpha=1.3$ is the sweet spot for the current baseline (D-ResNet-50). The increment or reduction of $\alpha$ will lead to performance deterioration.

\section{Discussion and Conclusion}
In this paper, we propose a novel contrastive loss for fully-supervised semantic segmentation, named Positive-Negative Equal contrastive loss, to introduce preferable distribution for hard pixels.
We redesign NCE loss function by increasing the importance of positive samples to cast off the stereotype of unsupervised learning. 
PNE loss achieves superior performance because more suitable positive samples can be provided in a supervised semantic segmentation way.
Besides, in order to optimize difficult samples precisely, the constructing strategy of individual anchor sets is applied in calculating contrastive loss. By individual optimizing based on the predicted class, the performance of the model has been further improved on the segmentation task.
Then, we reconstruct the negative sample pair and replace the confused embedding similarity with explicit label information to find difficult negative samples. 
Our method achieves promising results on various benchmarks for semantic segmentation tasks, which demonstrates the effectiveness of our method.

A possible limitation of this paper is that the source of positive and negative samples in this method is limited to one image. It may result in the process of contrastive learning for some categories not being carried out continuously and the performance boost not being large enough due to rarely providing enough positive/negative sample pairs. This situation will appear more likely in the category of small objects. 
In the previous method \cite{wang2021exploring, hu2021region, chen2020improved}, the memory bank solved this problem partly but introduced the problem of momentum update at the same time. 
However, it will corrupt the plug-and-play characteristic and sacrifice the convenience and compatibility of PNE loss in practice.
Therefore, how to solve the above limitations while retaining convenience is worth further research.

\section*{Acknowledgements}
This work was supported in part by the Natural Science Foundation of China under Grant 42201386, in part by the International Exchange Growth Program for Young Teachers of USTB under Grant QNXM20220033, and Scientific and Technological Innovation Foundation of Shunde Innovation School, USTB (BK20BE014).

\bibliographystyle{elsarticle-num} 
\bibliography{references}
\end{document}